\documentclass[letterpaper]{article} 
\usepackage[draft]{aaai2026}  
\usepackage{times}  
\usepackage{helvet}  
\usepackage{courier}  
\usepackage[hyphens]{url}  
\usepackage{graphicx} 
\usepackage{natbib}  
\usepackage{caption} 
\usepackage{algorithm}
\usepackage{algorithmic}

\usepackage[dvipsnames]{xcolor}
\usepackage{booktabs}
\usepackage{amsmath}

\usepackage{newfloat}
\usepackage{listings}
\DeclareCaptionStyle{ruled}{labelfont=normalfont,labelsep=colon,strut=off} 
\floatstyle{ruled}
\newfloat{listing}{tb}{lst}{}
\floatname{listing}{Listing}
\title{HAPO: Training Language Models to Reason Concisely via History-Aware Policy Optimization}
\author{
    Chengyu Huang\equalcontrib,
    Zhengxin Zhang\equalcontrib,
    Claire Cardie
}
\affiliations{
    Department of Computer Science, Cornell University\\

    ch2263@cornell.edu, zz865@cornell.edu, ctc9@cornell.edu
}

\usepackage{bibentry}

\begin{document}

\maketitle

\begin{abstract}
While scaling the length of responses at test-time has been shown to markedly improve the reasoning abilities and performance of large language models (LLMs), it often results in verbose outputs and increases inference cost. Prior approaches for efficient test-time scaling, typically using universal budget constraints or query-level length optimization, do not leverage historical information from previous encounters with the same problem during training. We hypothesize that this limits their ability to progressively make solutions more concise over time. To address this, we present History-Aware Policy Optimization (HAPO), which keeps track of a history state (e.g., the minimum length over previously generated correct responses) for each problem. HAPO employs a novel length reward function based on this history state to incentivize the discovery of correct solutions that are more concise than those previously found. Crucially, this reward structure avoids overly penalizing shorter incorrect responses with the goal of facilitating exploration towards more efficient solutions. By combining this length reward with a correctness reward, HAPO jointly optimizes for correctness and efficiency. We use HAPO to train DeepSeek-R1-Distill-Qwen-1.5B, DeepScaleR-1.5B-Preview, and Qwen-2.5-1.5B-Instruct, and evaluate HAPO on several math benchmarks that span various difficulty levels. Experiment results demonstrate
that HAPO effectively induces LLMs’ concise reasoning abilities, producing  length reductions of 33-59\% with accuracy drops of only 2-5\%.
\end{abstract}

\begin{links}
    \link{Code}{https://github.com/HCY123902/HAPO}
\end{links}

\section{Introduction}

Recent advances in large language models (LLMs) have highlighted the effectiveness of test-time scaling \cite{deepseek2025,open2024,zeng2025,deepscaler2025,deepcoder2025,qwq32b}, which enables LLMs to develop longer and more sophisticated reasoning behaviors such as self-reflection and verification, substantially improving their performance across a wide range of tasks. 
While generating long reasoning chains can significantly boost models' accuracy, it induces verbosity, increases inference cost, and incurs substantial computational and memory overhead due to the quadratic attention complexity and linear key-value (KV) cache growth that are inherent in transformer architectures \cite{vaswani2017}.
Notably, even for very simple problems like ``What is the answer to 2 plus 3?", these reasoning models tend to engage in unnecessarily lengthy reasoning and produce solutions that span hundreds of tokens when only a few steps would suffice \cite{kumar2025overthink,chen2024not,sui2025}. This inefficiency, often referred to as \textit{overthinking}, limits the practical deployment of reasoning models in real-world applications.

There have been initial explorations into limiting the generation length of reasoning models, either by enforcing a fixed, universal budget constraint to be applied to each question (i.e., universal budget forcing) \cite{fu2024efficiently, muennighoff2024, han2024, aggarwal2025, hou2025}, or by dynamically optimizing the reasoning length for each question \cite{luo2025,arora2025,yeo2025,she2025,qu2025} (i.e., query-level optimization). 
\begin{figure}
    \centering
    \includegraphics[width=\linewidth]{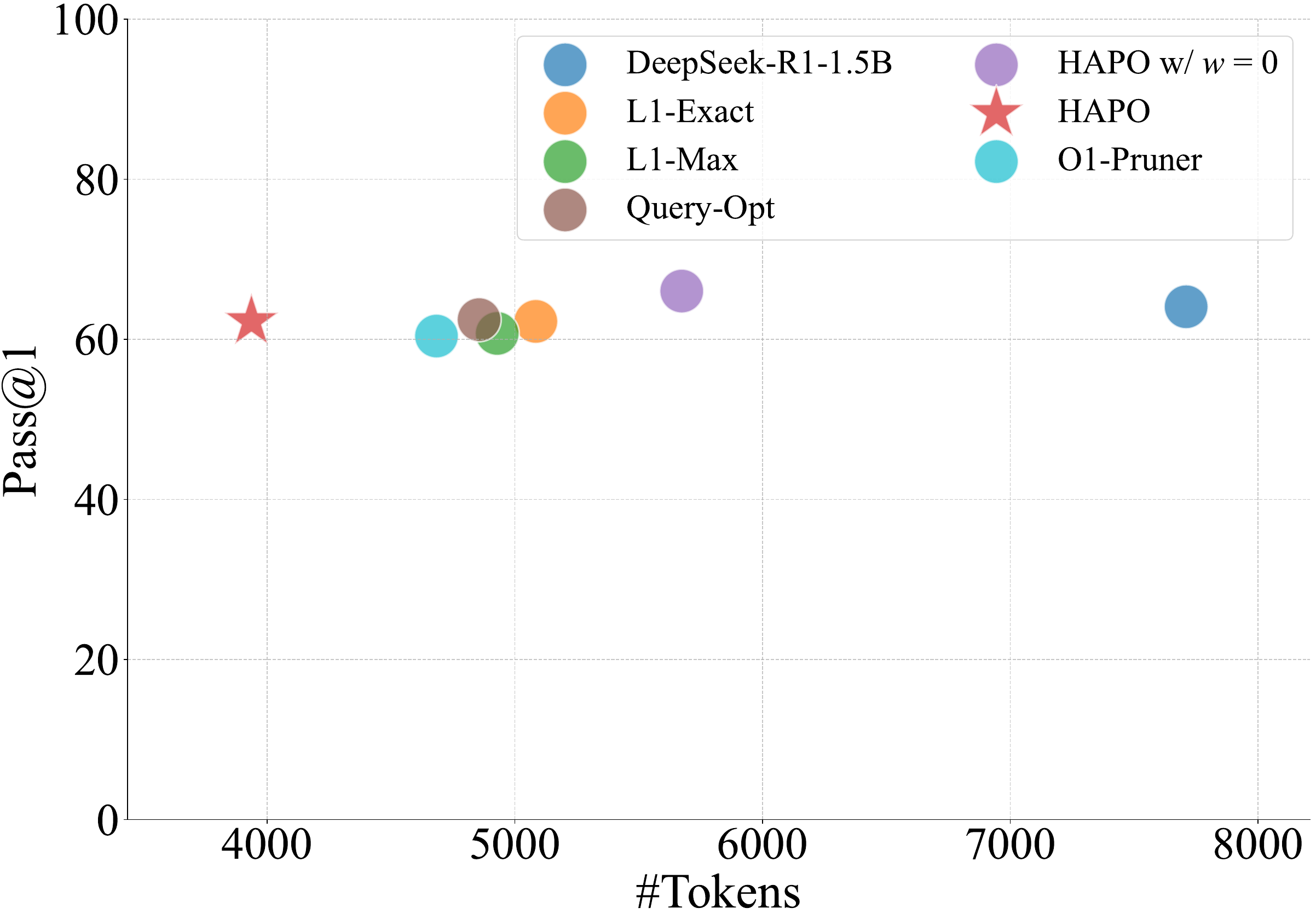}
    \caption{\label{fig:visualization_combined}Results averaged across GSM8K, MATH500, and AIME2024. Pass@1 is the average accuracy across multiple sampled responses per query; \#Tokens is the average number of tokens in the responses. Compared with the base model DeepSeek-R1-1.5B, HAPO significantly reduces response length while preserving accuracy, giving a better tradeoff than other baselines.
    }
\vspace{-\baselineskip}
\end{figure}
\textit{Universal budget forcing methods} intervene when a predefined token limit is reached. For reinforcement learning (RL) based methods, this usually results in a non-positive reward. For solutions that use supervised-finetuning (SFT) or few-shot prompting, it implies forcing a response length cutoff by inserting an \textsc{end-of-thinking} token. For example, S1 \cite{muennighoff2024} introduces a length control mechanism that injects special tokens (e.g., \textsc{Final Answer}) when responses are too long. Despite reducing output length with respect to a base model, such fixed token budgets cannot address the overthinking problem; instead they unnecessarily constrain exploratory reasoning on harder questions and still allow excessive token usage on simpler ones.

In contrast, \textit{query-level optimization methods} learn a different budget for each query. For example, \citet{arora2025} compare the lengths of the responses to the same query in each training step and reward shorter responses while penalizing longer ones.
Although these methods alleviate the overthinking problem to some extent, they cannot leverage historical information derived from previous encounters with the same query during training, i.e., across multiple training steps. This limits their ability to progressively make solutions more concise over time.
For example, suppose responses sampled initially were considerably longer than those sampled in a subsequent encounter with the same problem. 
We argue that in this case, the model has become more efficient and, therefore, should receive a positive length reward in the later encounter. 
Unfortunately,  existing approaches—including both universal budget forcing and query-level optimization methods—lack the mechanisms to utilize such historical signals.

In this paper, we propose \textbf{History-Aware Policy Optimization (HAPO)}. As illustrated in the left side of Figure~\ref{fig:example}, we introduce a history state $h_i$ that records the minimum length of the correct responses observed in previous encounters with problem $x_i$. We dynamically adjust the length reward (Figure~\ref{fig:example} right) based on $h_i$, assigning higher rewards to correct responses that are shorter than $h_i$. This encourages the model to generate correct solutions that are more concise than previously known ones, which is a different incentive from prior approaches. Additionally, unlike existing methods, we do not overly penalize short incorrect responses, under the assumption that they may represent exploratory attempts toward more concise solutions. HAPO jointly optimizes for both accuracy and efficiency by combining the length reward with an accuracy reward.


We use HAPO to train three open-source reasoning models, including DeepSeek-R1-Distill-Qwen-1.5B \cite{deepseek2025}, DeepScaleR-1.5B-Preview, and Qwen-2.5-1.5B-Instruct, and evaluate the models on math problems from three standard datasets---GSM8K \cite{cobbe2021}, MATH500 \cite{hendrycks2021}, and AIME2024.
%
Overall we find that 
HAPO effectively elicits LLMs' concise reasoning abilities: when averaged across the three benchmarks, HAPO produces a 49\% length reduction with only 2\% accuracy drop for DeepSeek-R1-Distill-Qwen-1.5B; a 33\% length reduction with 5\% accuracy drop for  DeepScaleR-1.5B-Preview; and a 59\% length reduction with 2\% accuracy drop for Qwen-2.5-1.5B-Instruct.
Using DeepSeek-R1-Distill-Qwen-1.5B as the base model, we compare HAPO with baselines from universal budget forcing (the L1 models \cite{aggarwal2025}) and query-level optimization (Query-Opt \cite{arora2025}) paradigms. Here, our results show
that HAPO enables a better overall length-correctness trade-off. Notably, HAPO uses 19\% fewer tokens than the most competitive baseline while achieving similar accuracy (see Figure~\ref{fig:visualization_combined}).

\section{Related Work}

\paragraph{Test-time scaling} Prior work \cite{wei2022, snell2025, wu2024} has shown that increasing test-time computation can improve the performance of large language models (LLMs), particularly in reasoning-intensive tasks such as math problem solving and code generation. \citet{wei2022} demonstrates that generating intermediate reasoning steps (i.e., chain-of-thought (CoT)) can effectively elicit the reasoning capabilities of LLMs, inspiring a wave of follow-up research. Many subsequent methods focus on parallel scaling, where multiple responses are sampled from the model and aggregated into a final answer, either through an external reward model or an internal voting mechanism. Representative approaches include Best-of-N sampling \cite{wu2024}, beam search \cite{snell2025}, majority voting \cite{wang2023} and its weighted variant \cite{li2023}, as well as tree-based search methods \cite{yao2023, xin2025, besta2024}.

Another line of work focuses on scaling the \textit{length} of the response rather than the \textit{number} of responses. It aims to enable LLMs to engage in more thorough reasoning and support self-correction and verification \cite{kumar2025}. With the success of OpenAI's o1 \cite{open2024} model, more recent efforts have been made along this line. Notable efforts include DeepSeek’s R1 models \cite{deepseek2025}, Qwen’s QwQ-series models, and other work that reproduces long CoT capabilities in smaller-scale models \cite{zeng2025, muennighoff2024}. A common feature of how most of these models are trained is the use of reinforcement learning (RL) algorithms \cite{schulam2017ppo, shao2024} and rule-based rewards. This training paradigm encourages models to generate increasingly long CoT in order to arrive at the correct answer.

\paragraph{Efficient reasoning}
While test-time scaling with long CoT significantly improves accuracy, it comes at the cost of computational inefficiency. In particular, reasoning models often produce verbose and unnecessary reasoning when solving simple problems—a phenomenon commonly referred to as \textit{overthinking} \cite{sui2025}. Empirically, we observe that reasoning models such as DeepSeek-R1-Distill-Qwen-1.5B can generate over 6K tokens when solving problems from the MATH dataset \cite{hendrycks2021}, whereas human-written solutions only have around 200 tokens on average.

A variety of methods have been proposed to address this issue. For example, some approaches \cite{geiping2025, hao2024, cheng2024, su2025, shen2025efficient, shen2025codi} replace discrete vocabulary tokens with latent tokens in the continuous embedding space to perform reasoning. Other methods remain in the vocabulary space and focus on prompt optimization or additional training. \citet{xu2025} designs a prompt that explicitly encourages conciseness at each reasoning step, while \citet{xia2025, han2024, shen2025dast} employ offline training algorithms to teach LLMs to generate concise yet correct responses. However, such offline training typically requires carefully curated supervision data, which can be expensive and labor-intensive to obtain.

In contrast, online RL training integrates length control directly within the reward signal. While the design of such length rewards varies across prior work, they generally fall into two categories: (1) Universal rewards, which compare response length against a predefined, universal budget and penalize responses exceeding this budget \cite{hou2025, aggarwal2025}, or utilize a universal penalty function \cite{yeo2025}; (2) Query-level rewards, which are computed according to LLMs' performance specific to each query. Many query-level rewards are comparison-based; some methods compare the lengths of responses to the same query within a batch, penalizing relatively longer ones \cite{kimi2025, arora2025}, while others compare the current response length against that of a reference model or ground truth \cite{luo2025, she2025}. We argue that universal rewards are suboptimal due to their lack of adaptivity, necessitating a manual specification of budget estimates or penalty functions. Although query-level rewards seem more promising, current designs need improvement. For instance, in-batch comparison may not achieve a global optimum (potentially rewarding long responses if all outputs in a batch are long), while comparison against a fixed reference can lead to rapid reward saturation once the model consistently outperforms the reference.

In this work, we instead follow the idea of utilizing past information from training history \cite{zhang2023, shinn2023, le2025}, to dynamically adjust the length reward per query, thereby providing the most up-to-date stimulus for the LLMs to compress their responses according to their own performance in the previous training steps.

\section{History-Aware Length Reward}
\label{sec:method}
In this section, we will first define our task setting and then describe our history-aware length reward.

\subsection{Task Setup}
\label{sec:ts}
Denote a $\theta$ parameterized LLM as $p_\theta$. Using the GRPO algorithm\footnote{We explore other RL approaches such as PPO \cite{schulam2017ppo} in the Appendix.} \cite{shao2024} we train $p_\theta$ on a dataset $\mathcal{D}=\{(x_i, a^*_i)\}_{i=1}^{N}$, where $x_i$ is a query 
(e.g., a math problem) and $a^*_i$ is its ground truth answer (e.g, a number). 
For each query $x_i$ in a batch, a set of responses $Y_{i} = \{y_0, y_1, ...\}$ is sampled from the current LLM $p_\theta$. 
Following the procedure in \cite{deepseek2025}, we extract a candidate final answer $a_i$ from each response $y_i$.
We train $p_\theta$ for multiple epochs; thus the model encounters each $x_i$ multiple times and we use $Y_{i}^{(j)}$  to refer to the set of responses generated by  $p_\theta$ in its $j$-th encounter with $x_i$. 
Computation of the reward associated with $x_i$ is described next.

\subsection{Reward Computation and History Update}
\label{sec:rchu}

The HAPO reward function for the $j$-th occurrence of $x_i$ is comprised of two parts, an accuracy reward, $ra_{i}^{(j)}$, and a length reward, $rl_{i}^{(j)}$.

\paragraph{Accuracy reward} We employ a \textit{binary accuracy reward} that compares the extracted answer $a_i$ against the ground truth $a^*_i$: $ra_{i}^{(j)} = 1$ if $a_{i}^{(j)} = a_{i}^*$, and equals $0$ otherwise.

However, models trained with only this rule-based correctness reward tend to generate verbose responses even for simple questions. As a result, we design a novel length reward that uses the training history to encourage conciseness while preserving correctness.
The motivation behind our length reward is that if $p_\theta$ gives a correct response $y_i$ to query $x_i$, then in subsequent training if $p_\theta$ encounters $x_i$ again, it should be positively rewarded if the new response is still correct but is shorter than $y_i$; otherwise (i.e., the new response to $x_i$ is longer than $y_i$ or is incorrect), it needs to be penalized.
\begin{figure*}[h]
    \centering
    \includegraphics[width=\linewidth]{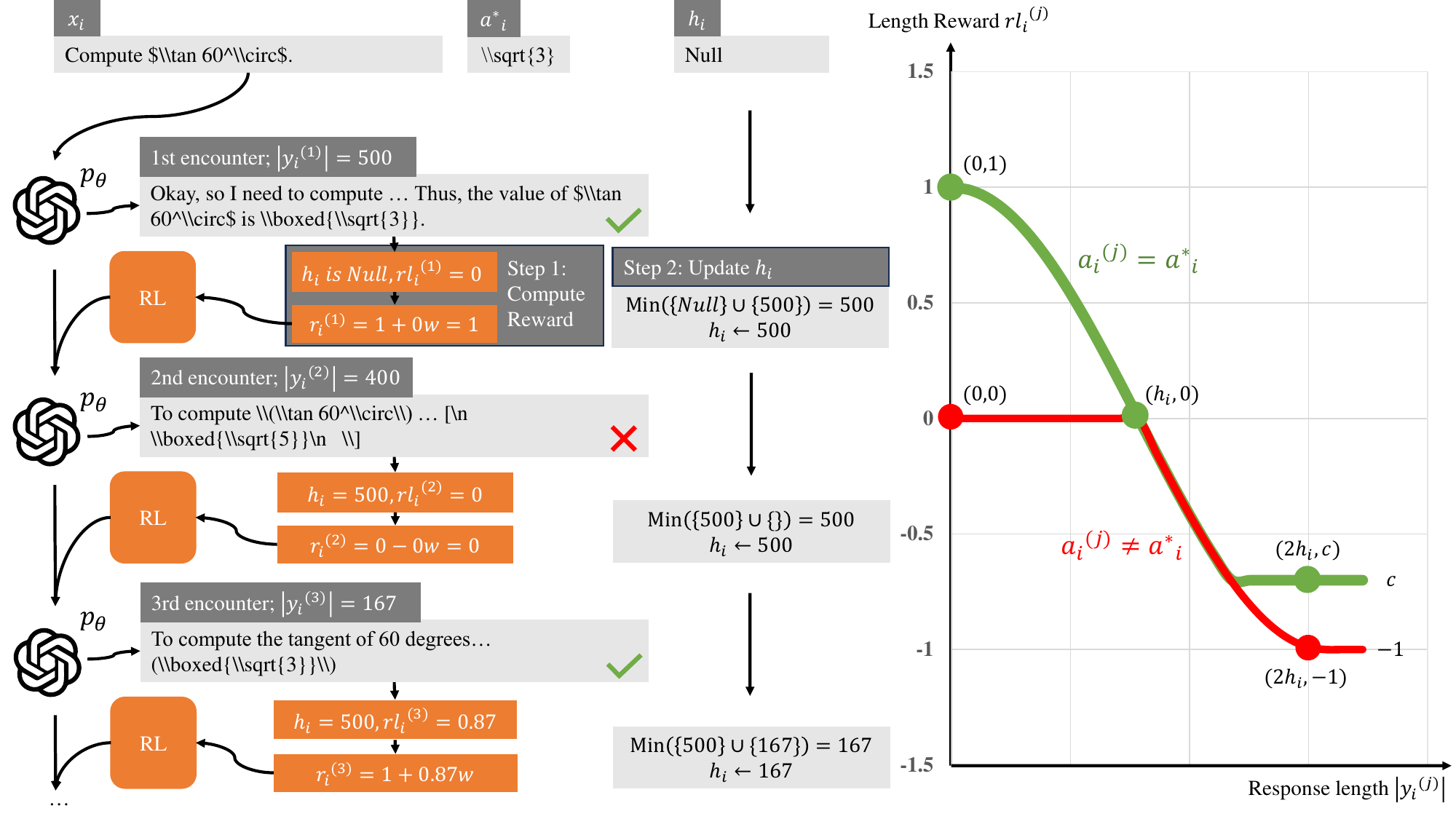}
    \caption{
\textbf{Left}: 
Illustration of the HAPO reward computation and history state ($h_i$) update over three encounters with problem $x_i$, assuming one response is sampled per encounter ($|Y_{i}^{(j)}|=1$). Initially ($j=1$, $h_i=\texttt{Null}$). The first correct response $y_{i}^{(1)}$ (length 500) receives a zero length reward ($rl=0$) and an overall reward of 1 ($r=1$), and updates $h_i$ to 500. 
In the second encounter ($j=2$), an incorrect response $y_{i}^{(2)}$ (length 400) receives $rl=0$ and $r=0$, and does not update $h_i$. 
In the third encounter ($j=3$), a correct response $y_{i}^{(3)}$ (length 167), being shorter than the current $h_i=500$, receives a positive length reward ($rl=0.87$) and overall reward ($r \ge 1$), and updates $h_i$ to 167.
\textbf{Right}: 
The HAPO length reward function $rl_i^{(j)}$ plotted against response length $|y_i^{(j)}|$. \textbf{\textcolor{ForestGreen}{Green curve}}: reward for correct responses ($a_{i}^{(j)}=a_i^*$), incentivizing lengths shorter than $h_i$; \textbf{\textcolor{red}{Red curve}}: reward for incorrect responses ($a_{i}^{(j)} \neq a_i^*$). Both curves are centered at $(h_i, 0)$. The initial case where $h_i$ is \texttt{Null} is omitted for visual clarity.
}
    \label{fig:example}
\vspace{-0.5\baselineskip}
\end{figure*}
To compute the length reward, we maintain a \textit{history reference length} $h_i$ for each query $x_i$ in the training set. 
$h_i$ is initialized as \texttt{Null} = \textsc{MaxLength} (i.e., set to max length for update but treated as \texttt{Null} for reward computation) 
and is updated 
whenever a shorter correct response for $x_i$ is produced during training. 
Details of the length reward computation
are described below and the overall procedure is illustrated in Figure~\ref{fig:example} (left). 

\paragraph{Length reward design} The length reward will be computed for \textit{each} response $y_{i}^{(j)} \in Y_{i}^{(j)}$. At a high level, we desire a length 
reward with the following characteristics. (1) If  $|y_{i}^{(j)}|<h_{i}$ and $a_{i}^{(j)}$ is correct, we should assign a positive reward, since $y_{i}^{(j)}$ is a more concise solution than any previous correct one. (2) If $|y_{i}^{(j)}|>h_{i}$ and $y_{i}^{(j)}$ is correct, then we should assign a negative reward. This is because $y_{i}^{(j)}$ is suboptimal with the presence of a shorter correct response. (3) If $y_{i}^{(j)}$ is incorrect but $|y_{i}^{(j)}|<h_{i}$, we hypothesize that $p_\theta$ is exploring a \textit{potentially} correct and shorter response and so assign a neutral reward of 0. (4) If $|y_{i}^{(j)}|>h_{i}$ and $y_{i}^{(j)}$ is incorrect, then, similar to (2), we want a negative reward since $y_{i}^{(j)}$ is suboptimal. 
(5) In contrast to prior work \cite{yeo2025, shen2025dast}, we also want the reward to decrease smoothly as response length increases, even for incorrect responses, as long as $h_i$ is not \texttt{Null}.  
(6) The length reward should ensure that any correct response has an overall reward $r_{i}^{(j)}$ that is strictly higher than any incorrect one.
(7) Lastly, if the reference $h_{i}$ is still \texttt{Null}, then this means $p_\theta$ has not generated any correct response in the training history, and therefore we give the current response a reward of 0 since no comparison can be made.

Figure~\ref{fig:example} (right) depicts our length reward function that ranges from -1 to 1, rewards correct responses shorter than $h_{i}$, penalizes all responses longer than $h_{i}$, and smoothly decreases as $|y_{i}^{(j)}|$ increases.  It relies on the cosine function in $f$ to meet characteristic (5) as well as a clipping cutoff $c\in(-1,0)$ for the $a_{i}^{(j)}=a^*_{i}$ case to meet characteristic (6): 
\begin{align*}
    rl_{i}^{(j)}=
    \begin{cases}
        max\left(f(|y_{i}^{(j)}|, h_{i}), c\right) & \text{if } a_{i}^{(j)}=a^*_{i}\\
        min\left(f(|y_{i}^{(j)}|, h_{i}),0\right) & \text{if } a_{i}^{(j)}\neq a^*_{i}\\
        0 & \text{if } h_{i} \text{ is \texttt{Null}}
    \end{cases}
\end{align*}
where $f(|y_{i}^{(j)}|, h_{i}) = cos\left(min(\frac{\pi}{2}\frac{|y_{i}^{(j)}|}{h_{i}}, \pi)\right)$. 

\paragraph{Final reward function} The final reward combines the length reward with the accuracy reward:
\begin{align*}
    r_{i}^{(j)}=1(a_{i}^{(j)}=a^*_i) + w\cdot rl_{i}^{(j)}
\end{align*}
$w$ is a hyperparameter within the range $[0, 1]$ that controls the weight of the length component of the overall reward.

\paragraph{History state update} After computing the reward for each response in $Y_{i}^{(j)}$, we update the history state $h_i$. Specifically, we extract the lengths of all correct responses in $Y_{i}^{(j)}$ as $L_{i}^{(j)}=\{|y_{i}^{(j)}|\vert y_{i}^{(j)}\in Y_{i}^{(j)} \wedge a_{i}^{(j)}=a^*_i\}$. We then update $h_i$ using an aggregation function $aggre$ (i.e., $h_i\leftarrow aggre(\{h_i\}\cup L_{i}^{(j)})$). In this work, we set $aggre$ to the minimum function, so $h_i$ tracks the shortest correct response observed thus far. Other aggregation choices, such as taking the mean or median, are also possible.

\section{Experiment Setup}
\label{sec:setup}

\paragraph{Training Details} 
\label{sc:td}
We train our models on 2,000 examples from the DeepScaleR-Preview-Dataset \cite{deepscaler2025} and validate the models on another 500. The original dataset contains 40,000 math problems drawn from a variety of sources, spanning a difficulty spectrum from grade-school level (e.g., GSM8K \cite{cobbe2021}) to competitive mathematics (e.g., AMC, AIME). For the base models, we experiment with DeepSeek-R1-Distill-Qwen-1.5B (\textbf{DeepSeek-R1-1.5B} for short), DeepScaleR-1.5B-Preview (\textbf{DeepScaleR-1.5B} for short), and Qwen-2.5-1.5B-Instruct (\textbf{Qwen-2.5-1.5B-Inst} for short).

Both DeepSeek-R1-1.5B and DeepScaleR-1.5B are reasoning models, where the latter is an improved version of the former, with higher accuracy and shorter responses on reasoning tasks. In contrast, Qwen-2.5-1.5B-Inst is not a reasoning model, and is not trained specifically to generate long reasoning traces to solve complex questions. To apply HAPO to Qwen-2.5-1.5B-Inst, we curate an easier training set that comprises 2,000 examples from the train split of MATH dataset. This is to ensure that the model can still generate correct answers and receive positive length rewards for a reasonable number of queries during training. All models are trained for 5 epochs using the GRPO algorithm with HAPO hyperparameters set to $w = 1.0$ and $c = -0.7$.
More training details are in the Appendix.

\paragraph{Evaluation}
\label{sc:eval}
We evaluate the trained models on GSM8K \cite{cobbe2021}, MATH500 \cite{hendrycks2021}, and AIME2024, which represent grade-school, intermediate, and competition-level math problems, respectively. Following \citet{deepseek2025}, we use a sampling temperature of 0.6, top-$p$ of 0.95, and a maximum context length of 32,768 tokens. For GSM8K and MATH, we generate 4 responses per query, and for AIME2024, we generate 32 responses per query. As in prior work, we report \textbf{Pass@1} (i.e., accuracy averaged across multiple sampled responses) and the average \textbf{\#Tokens} in the responses for each benchmark.

\paragraph{Baselines}
\label{sc:baseline}
We compare our method against two baselines: (1) the untrained base model and (2) an ablation baseline, denoted as \textbf{HAPO w/ $w = 0$}. In the ablation baseline, the length reward $rl$ is removed from the total reward $r_i^{(j)}$ by setting its weight $w$ to 0, such that the model is trained to optimize solely for correctness. This setup allows us to isolate the impact of length reward in HAPO.

Since many existing approaches build upon DeepSeek-R1-1.5B and have released code or model checkpoints, we additionally compare HAPO in this setting against several representative methods from two paradigms: universal budget forcing and query-level optimization.

For universal budget forcing, we compare against \textbf{L1-Exact} and \textbf{L1-Max} \cite{aggarwal2025}. Both methods employ GRPO with a token budget penalty alongside the accuracy reward. \textbf{L1-Exact} trains models to produce responses with exactly the target budget, while \textbf{L1-Max} treats the budget as an upper bound. To ensure a fair comparison—since the original models were trained on a significantly larger dataset—we rerun these methods using our own training data. For each benchmark, the L1-Exact budget is set to the average response length of HAPO, while L1-Max is given twice that average. Although exploring multiple token budgets for L1 baselines may improve their length-correctness trade-offs, we avoid this to limit compute and human interventions, and maintain fairness.

For query-level optimization, we include the method proposed by \citet{arora2025}, denoted as \textbf{Query-Opt}. This method compares the lengths of responses for the same query in each training batch, rewarding relatively shorter responses and penalizing longer ones. Additionally, we evaluate \textbf{O1-Pruner} \cite{luo2025}, which uses the base model's responses to each query as a reference, to compute length and correctness rewards.


\section{Results and Analysis}
\label{sec:res}

We first present the main results 
on three models: DeepSeek-R1-1.5B, DeepScaleR-1.5B, and Qwen-2.5-1.5B-Inst. We then focus on DeepSeek-R1-1.5B for further experiments and analysis, including comparisons with prior work, 
 a case study on MATH500, 
 and an inspection of training dynamics.
 Finally, we examine the impact of training set size on the model's performance. 

\subsection{Main Results}
\label{sec:res_main}


\begin{table*}[h]
  \centering
  \resizebox{\textwidth}{!}{
  \begin{tabular}{lcccccccc}
    \toprule
    & \multicolumn{2}{c}{GSM8K} &  \multicolumn{2}{c}{MATH500} & \multicolumn{2}{c}{AIME2024} & \multicolumn{2}{c}{Average} \\
    \cmidrule(r){2-9} 
    Models     & Pass@1 \textcolor{red}{$\uparrow$}  & \#Tokens \textcolor{red}{$\downarrow$} & Pass@1 \textcolor{red}{$\uparrow$}  & \#Tokens \textcolor{red}{$\downarrow$}  & Pass@1 \textcolor{red}{$\uparrow$}  & \#Tokens \textcolor{red}{$\downarrow$} & Pass@1 \textcolor{red}{$\uparrow$}  & \#Tokens \textcolor{red}{$\downarrow$} \\
    \midrule
    DeepSeek-R1-1.5B & 83.65  & 2066  & 79.15   & 6449  & 29.38      & 14615  & 64.06 & 7710 \\
    HAPO w/ $w=0$ & \textbf{84.40}  & 1381          & \textbf{82.15}   & 4939          & \textbf{31.15}  & 10701  & \textbf{66.02} & 5674 \\
    HAPO          & 79.08           & \textbf{661}   & 81.05           & \textbf{2978} & 26.56  & \textbf{8171}  & 62.23 & \textbf{3937} \\
    \midrule
    DeepScaleR-1.5B & 86.55  & 1702  & \textbf{88.05}  & 3765  & 37.92  & 8826  & 70.84 & 4764 \\
    HAPO w/ $w=0$ & \textbf{86.81}  & 1509  & 88.00  & 3462  & \textbf{40.10}  & 8094  & \textbf{71.64} & 4355 \\
    HAPO & 83.59  & \textbf{1122}  & 84.20  & \textbf{2681}  & 31.15  & \textbf{5813}  & 66.31 & \textbf{3205} \\
    \midrule
    Qwen-2.5-1.5B-Inst & 70.43  & 322  & 53.50 & 799  & \textbf{3.54}  & 3923 & 42.49 & 1681 \\
    HAPO w/ $w=0$ & \textbf{72.95}  & 301  & \textbf{54.75}  & 589  & 2.71  & 1643 & \textbf{43.47} & 844 \\
    HAPO & 67.90  & \textbf{200}  & 52.25  & \textbf{390}  & 2.81  & \textbf{1493} & 40.99 & \textbf{694} \\
    \bottomrule
  \end{tabular}}
  \caption{\label{tab:res_main} Main results on GSM8K, MATH500, and AIME2024.}
\end{table*}

\textbf{HAPO achieves substantial length reduction while preserving most of the accuracy.} As shown in Table~\ref{tab:res_main}, on DeepSeek-R1-1.5B, compared with the base model, HAPO reduces response length by 68\%, 54\%, and 44\% on GSM8K, MATH500, and AIME2024, respectively. The corresponding changes in accuracy are a 5\% drop, a 2\% gain, and a 3\% drop. Overall, HAPO achieves a 49\% reduction in response length with only a 2\% decrease in accuracy.

On DeepScaleR-1.5B, HAPO reduces response length by 34\%, 29\%, and 34\% across the three benchmarks, with accuracy drops of 3\%, 4\%, and 7\%, respectively. The gains here are less pronounced, likely because DeepScaleR-1.5B has already been fine-tuned to reduce response length from the original DeepSeek-R1-1.5B model. This can leave less room for further length compression.

On Qwen-2.5-1.5B-Inst, HAPO again proves effective, reducing response length by 38\%, 51\%, and 62\%, while keeping accuracy drops within 3\%, 1\%, and 1\% on the three benchmarks. These suggest that HAPO is broadly effective, even for models not originally optimized for reasoning.

Consistent with the findings in \cite{arora2025}, we observe a reduction in token usage even when the model is trained solely with the correctness reward (i.e., HAPO w/ $w = 0$). We hypothesize that this may be due to the relatively low difficulty of questions in the training set, though further investigation is needed to confirm the underlying cause. Compared to this baseline, HAPO produces responses that are, on average, 31\% shorter and 4\% less accurate on DeepSeek-R1-1.5B across all benchmarks. The corresponding reductions for DeepScaleR-1.5B and Qwen-2.5-1.5B-Inst are (26\%, 5\%) and (18\%, 2\%), respectively. These suggest that HAPO achieves a more favorable length-accuracy trade-off than optimizing for correctness alone.

\subsection{Comparison with Prior Work}
\label{sec:prior_work_comparsion}

\begin{table*}[h]
  \centering
  \resizebox{\textwidth}{!}{
  \begin{tabular}{lcccccccc}
    \toprule
    & \multicolumn{2}{c}{GSM8K} &  \multicolumn{2}{c}{MATH500} & \multicolumn{2}{c}{AIME2024} & \multicolumn{2}{c}{Average}                \\
    \cmidrule(r){2-9} 
    Models     & Pass@1 \textcolor{red}{$\uparrow$}  & \#Tokens \textcolor{red}{$\downarrow$} & Pass@1 \textcolor{red}{$\uparrow$}  & \#Tokens \textcolor{red}{$\downarrow$}  & Pass@1 \textcolor{red}{$\uparrow$}  & \#Tokens \textcolor{red}{$\downarrow$} & Pass@1 \textcolor{red}{$\uparrow$}  & \#Tokens \textcolor{red}{$\downarrow$} \\
    \midrule
    DeepSeek-R1-1.5B & \underline{83.65}  & 2066  & 79.15   & 6449  & 29.38      & 14615  & \underline{64.06} & 7710 \\
    L1-Exact (U)     & 79.31           & 963           & \textbf{82.70}  & 3366          & 24.69   &  10927 & 62.23 & 5085 \\
    L1-Max (U)       & 80.33           & 1239          & 81.55           & 3404          & 20.30  & \underline{10144}  & 60.73 & 4929 \\
    Query-Opt (Q)     & 79.22           & \textbf{478}  & 81.45           & 3284  & 26.67  & 10805  & 62.45 & 4856 \\
    O1-Pruner (Q)     & 76.99           & \underline{497}  & 71.80           & \underline{3264}  & \textbf{32.42}  & 10290  & 60.40 & \underline{4684} \\
    HAPO w/ $w=0$ & \textbf{84.40}  & 1381          & \underline{82.15}   & 4939          & \underline{31.15}  & 10701  & \textbf{66.02} & 5674 \\
    HAPO          & 79.08           & 661   & 81.05           & \textbf{2978} & 26.56  & \textbf{8171}  & 62.23 & \textbf{3937} \\
    \bottomrule
  \end{tabular}}
  \caption{\label{tab:pw_comparison} Comparison with prior work on DeepSeek-R1-1.5B. (U) denotes universal budget forcing methods; (Q) denotes query-level optimization methods.}
\end{table*}


\textbf{HAPO achieves a superior length-accuracy trade-off compared with prior work.} In Table~\ref{tab:pw_comparison}, we compare HAPO against L1-Exact, L1-Max, Query-Opt, and O1-Pruner.

On MATH500, HAPO produces the shortest responses, using 9\% fewer tokens than the second place. It attains an accuracy that is similar to the prior methods, except for O1-Pruner, whose accuracy is much lower. On AIME2024, HAPO again gives the most concise responses, using 19\% fewer tokens than the second place. In terms of accuracy, it is superior or on par with prior methods, except that O1-Pruner's accuracy stands out higher. On GSM8K, HAPO is less effective than Query-Opt and generates longer responses, though it still outperforms the other methods.

On average, HAPO’s responses are shorter than L1-Exact, L1-Max, Query-Opt, and O1-Pruner by 23\%, 20\%, 19\%, and 16\% respectively, while maintaining similar accuracy.

\subsection{Case Study on MATH500}
\label{sec:case_study}

\begin{figure}
\vspace{-.8\baselineskip}
\centering
\includegraphics[width=\linewidth]{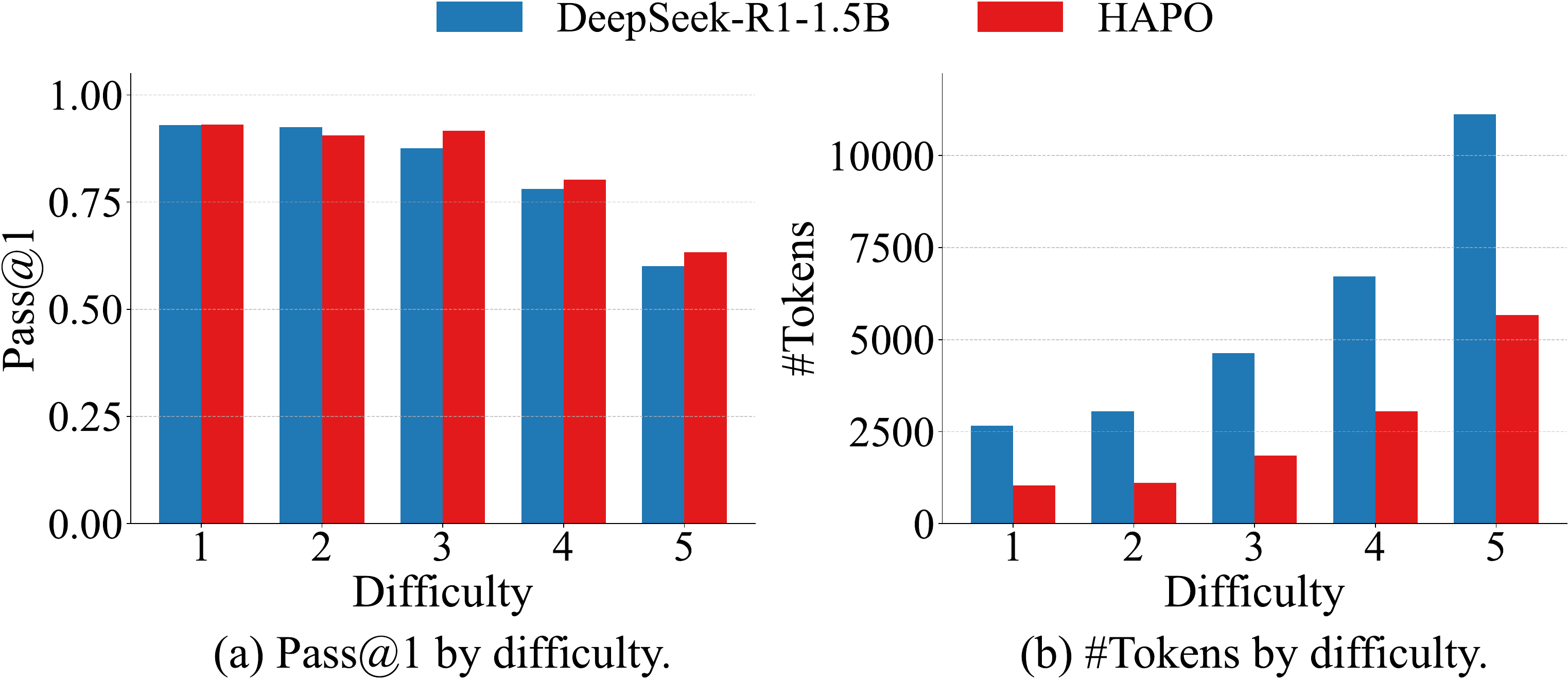}
\caption{\label{fig:case_study} Comparison between HAPO and DeepSeek-R1-1.5B in terms of Pass@1 and \#Tokens across questions of varying difficulty levels.}
\vspace{-\baselineskip}
\end{figure}
While Table~\ref{tab:pw_comparison} provides a high-level overview of HAPO’s performance across benchmarks, we conduct a more fine-grained case study on MATH500 to examine how HAPO performs across questions of varying difficulty levels. We obtain the difficulty labels from the original MATH dataset, which range from level 1 (easiest) to level 5 (hardest), and analyze Pass@1 accuracy and response length at each difficulty level. As shown in Figure~\ref{fig:case_study}, on easier questions (levels 1 and 2), HAPO largely preserves accuracy while reducing response length by 61\% and 64\%, respectively. On more challenging questions (levels 3 to 5), HAPO slightly improves accuracy while still achieving a response length reduction of at least 50\%. These results indicate that \textbf{HAPO consistently reduces response length while maintaining accuracy across varying levels of question difficulties}.


\subsection{Results on Out of Domain Benchmarks}
\label{sec:res_ood}

\begin{table*}
\vspace{-\baselineskip}
\centering
\begin{tabular}{lcccccc}
    \toprule
    & \multicolumn{2}{c}{GPQA} &  \multicolumn{2}{c}{CodeBench} & \multicolumn{2}{c}{Average} \\
    \cmidrule(r){2-7} 
    Models     & Pass@1 \textcolor{red}{$\uparrow$}  & \#Tokens \textcolor{red}{$\downarrow$} & Pass@1 \textcolor{red}{$\uparrow$}  & \#Tokens \textcolor{red}{$\downarrow$} & Pass@1 \textcolor{red}{$\uparrow$}  & \#Tokens \textcolor{red}{$\downarrow$} \\
    \midrule
    DeepSeek-R1-1.5B & 34.34 & 10296 & 16.42   & 15070 & 25.38 & 12683 \\
    L1-Exact (U) &\textbf{36.70} & \underline{6658} & 16.23 & 10965 & \textbf{26.47} & 8812 \\
    L1-Max (U) &33.43 & \textbf{6168} & 16.01 & \underline{10578}  & 24.72 & \textbf{8373} \\
    Query-Opt (Q) & \underline{35.86} & 8228 & \textbf{17.05} & 11456 & \underline{26.46} & 9842 \\
    O1-Pruner (Q) & 34.85 & 7944 & 16.34 & 10609 & 25.60 & 9277 \\
    HAPO w/ $w=0$ & 33.33 & 8611 & 16.67 & 11098 & 25.00 & 9855 \\
    HAPO & 32.32 & 7563 & \underline{16.84} & \textbf{9933} & 24.58 & \underline{8748} \\
    \bottomrule
  \end{tabular}
  \caption{\label{tab:res_ood} Results on GPQA and LiveCodeBench.}
\vspace{-.8\baselineskip}
\end{table*}
We evaluate HAPO on two out-of-domain benchmarks to assess its generalizability. GPQA \cite{rein2024gpqa} comprises multiple-choice questions on biology, physics, and chemistry, while LiveCodeBench \cite{jain2025} includes a variety of coding-related problems.

\begin{table*}[t]
  \centering
  \resizebox{\textwidth}{!}{
  \begin{tabular}{lcccccccc}
    \toprule
    & \multicolumn{2}{c}{GSM8K} &  \multicolumn{2}{c}{MATH500} & \multicolumn{2}{c}{AIME2024} & \multicolumn{2}{c}{Average} \\
    \cmidrule(r){2-9} 
    Models     & Pass@1 \textcolor{red}{$\uparrow$}  & \#Tokens \textcolor{red}{$\downarrow$} & Pass@1 \textcolor{red}{$\uparrow$}  & \#Tokens \textcolor{red}{$\downarrow$}  & Pass@1 \textcolor{red}{$\uparrow$}  & \#Tokens \textcolor{red}{$\downarrow$} & Pass@1 \textcolor{red}{$\uparrow$}  & \#Tokens \textcolor{red}{$\downarrow$} \\
    \midrule
    DeepSeek-R1-1.5B & 83.65  & 2066  & 79.15   & 6449  & 29.38  & 14615 & 64.06 & 7710 \\
    HAPO ($|\mathcal{D}|=2$k) & 79.08  & 661  & 81.05  & 2978  & 26.56  & 8171 & 62.23 & 3937 \\
    ~~w/ $|\mathcal{D}|=1k$ & 82.70 & 1108 & 83.55 & 3463 & 25.00 & 10408 & 63.75 & 4993 \\
    ~~w/ $|\mathcal{D}|=500$ & 84.06  & 1371  & 82.05  & 5194  & 29.90  & 10700 & 65.34 & 5755 \\
    ~~w/ $|\mathcal{D}|=100$ & 83.89  & 1785 & 80.80  & 5887  & 30.94 & 13518 & 65.21 & 7063 \\
    \bottomrule
  \end{tabular}}
  \caption{\label{tab:training_set_size} Pass@1 (accuracy) and \#Tokens (response length) when training on smaller datasets.}
\end{table*}

As shown in Table~\ref{tab:res_ood}, \textbf{training on math problems with HAPO enables the model to reason more concisely in other domains, albeit to a lesser extent}. On GPQA, HAPO reduces response length by 27\% with a 2\% drop in accuracy. On LiveCodeBench, it achieves a 34\% reduction in length while slightly improving accuracy.

Compared with prior work, HAPO achieves the best length-correctness tradeoffs on LiveCodeBench, with the shortest responses and similar accuracy. On GPQA, the results are mixed. HAPO produces shorter responses but lower accuracy than Query-Opt and O1-Pruner, and it underperforms the L1 methods. Further investigation into these domain-specific trade-offs is left for future work.

\subsection{Training Dynamics}
\label{sec:training_dynamics}

$h_i$ is a key component in HAPO. We inspect how the average response length $|y_i|$ and historical minimum length $h_i$ change as training progresses.

\begin{figure}[h]
\centering
\includegraphics[width=\linewidth]{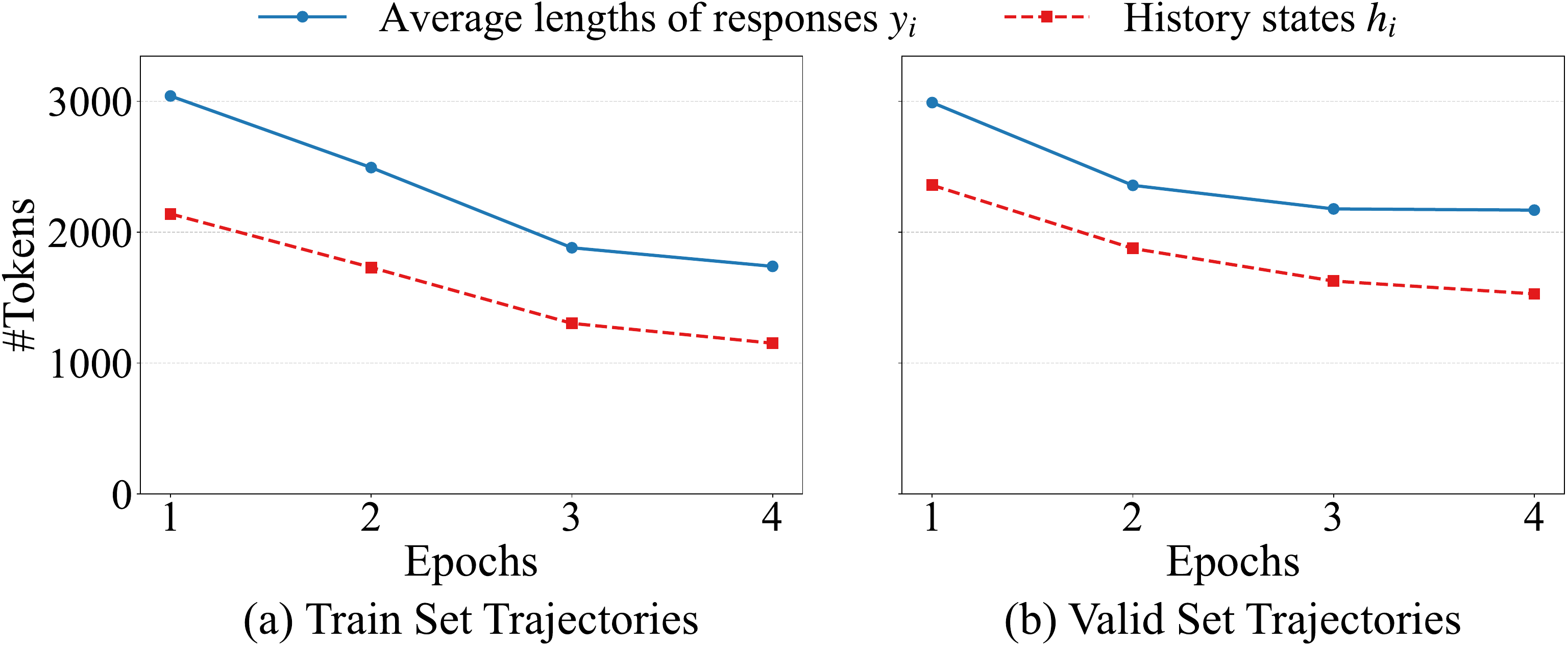}
\caption{\label{fig:training_dynamics} Trajectories of average response length $|y_i|$ and $h_i$ on the training set (left) and validation set (right), computed at the \textit{end} of each epoch. For the training set, the value of $|y_i|$ at each epoch is obtained by averaging the lengths of responses sampled from $p_\theta$ during that epoch across the entire training set. Similarly, $h_i$ is averaged over all training examples, excluding \texttt{Null} values. Note that $h_i$ is measured \textit{after} the update; that is, in epoch $j$, $h_i = \mathrm{aggre}({L_i^{(1)}, \cdots, L_i^{(j)}})$, which includes the latest length $L_i^{(j)}$.}
\vspace{-\baselineskip}
\end{figure}

On the training set (Figure~\ref{fig:training_dynamics}, left), both $|y_i|$ and $h_i$ steadily decrease as training progresses. In fact, the reductions in $|y_i|$ and $h_i$ reinforce each other: a decrease in average response length increases the likelihood of encountering an unprecedentedly short response, which in turn lowers $h_i$. A lower $h_i$ then acts as a stronger incentive for the model to produce even shorter responses in the next epoch, further reducing $|y_i|$. This mutually reinforcing trend is also observed on the validation set (Figure~\ref{fig:training_dynamics}, right). However, the average response length curve begins to plateau around epoch 4, suggesting potential overfitting with continued training.

\subsection{Training on Smaller Datasets for More Epochs}
\label{sec:training_on_smaller_datasets}

Prior works such as \cite{fatemi2025concise} suggest that training LLMs on a small set of occasionally solvable examples can effectively reduce their response length. Inspired by this, we check whether reducing the size of our training set will yield similar or even better performance. In particular, we reduce the training set size from the original 2,000 to 1,000, 500, and 100, respectively. In each setting, we train the model for 10 epochs to ensure a sufficient number of iterations while preventing overfitting. We keep other hyperparameters the same. Table~\ref{tab:training_set_size} shows the results.

Compared with the original setting, training on fewer examples leads to higher accuracy but longer responses. The overall performance becomes much closer to that of the base model. This is expected: with a smaller training set, the model is exposed to a more limited range of question types, increasing the likelihood of encountering unfamiliar questions during evaluation. As discussed previously, the concise reasoning abilities learned by HAPO do generalize to unseen questions, but to a limited extent. Consequently, the model deviates less from the base model.


\section{Conclusion}

In conclusion, we propose History-Aware Policy Optimization (HAPO), a training method that enables large language models to reason more concisely by leveraging historical information from past interactions. HAPO encourages models to iteratively discover correct responses that are more concise than those previously found. 
Experimental results demonstrate that HAPO significantly reduces response length while maintaining accuracy, achieving a more favorable length-correctness trade-off than prior approaches. 
Through the use of training history, HAPO opens new directions for building more adaptive, efficient, and capable models in reasoning tasks.

\bibliography{aaai2026}

\clearpage


\section{Visualization of Results for DeepSeek-R1-1.5B}

\begin{figure*}[ht]
    \centering
    \includegraphics[width=\linewidth]{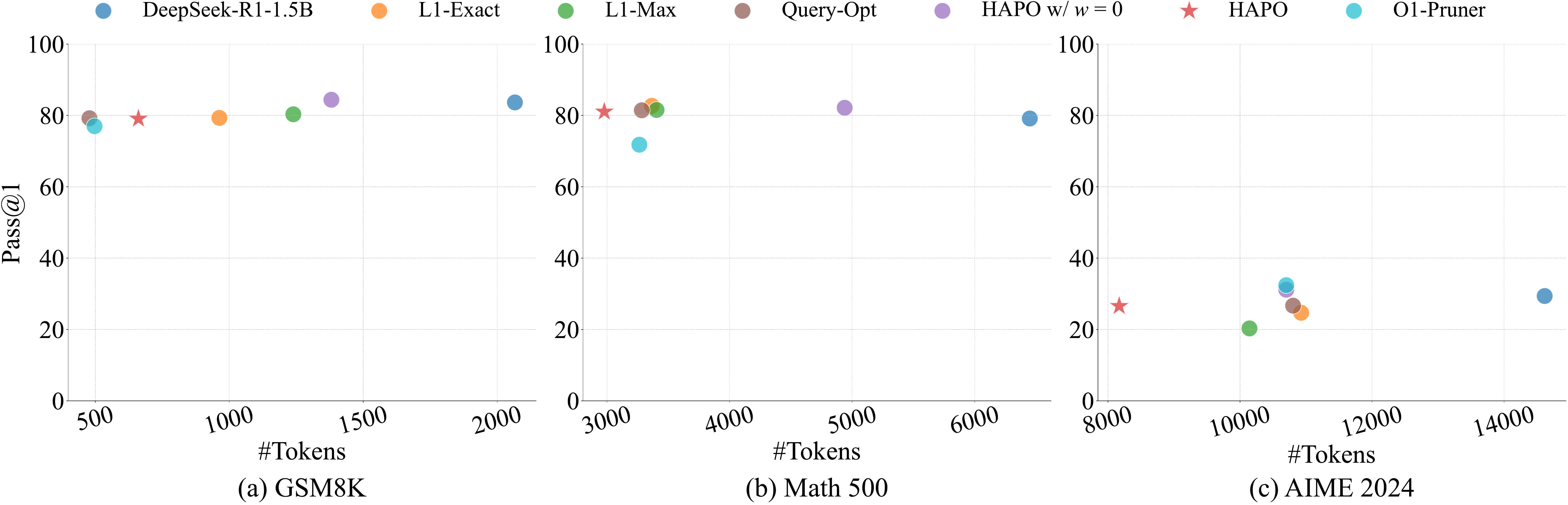}
    \caption{\label{fig:visualization} Visualization of results on GSM8K, MATH500, and AIME2024, with DeepSeek-R1-1.5B as the base model.
    }
\end{figure*}

Figure~\ref{fig:visualization} presents a comparison of HAPO and various baselines using DeepSeek-R1-1.5B as the base model. Compared with the base model, HAPO reduces response length by 68\%, 54\%, and 44\% on GSM8K, MATH500, and AIME2024, respectively, with corresponding accuracy changes of -5\%, +2\%, and -3\%. While HAPO slightly underperforms the Query-Opt baseline on GSM8K, the overall results indicate that it achieves a more favorable trade-off between response length and correctness than existing baselines.

\section{Reward Design}

Our reward design ensures that correct responses always receive strictly higher overall rewards than incorrect responses. In particular, the overall reward of any correct response is within the range $[1+wc, 1+w]$, where the lower bound $1+wc$ is achieved when $|y_i|\geq 2h_i$. Note than $c\in(-1,0)$ and $w\in[0,1]$, so $cw>-1$ and $1+wc>0$. The overall reward of any incorrect response is within the range $[-w,0]$, where the upper bound $0$ is achieved when $|y_i|\leq h_i$.

\section{Training and Reproduction Details}
\label{app:training_details}

We train our models using the Adam optimizer with a learning rate of 1e-6 and a linear learning rate scheduler. The batch size is set to 2, with a gradient accumulation step of 4. During training, we sample responses from $p_\theta$ for each query in a batch using a temperature of 0.7. The best checkpoint is selected based on the performance on a separate set of 500 examples from the DeepScaleR-Preview-Dataset. All experiments are conducted using 4 A6000 GPUs and 2 A100 GPUs.

To improve the efficiency of tensor computation, we initialize $h_i$ with a large value, \textsc{MaxLength} = 100,000, in place of \texttt{Null}, and treat it as an alias for \texttt{Null} in subsequent reward computations and history state updates. 

In both training and evaluation, we use the following system prompt
\begin{center}
    \texttt{Please reason step by step, and put your final answer within \textbackslash boxed\{\}.}
\end{center}
and set the user message to the question text to get the model's response.

\section{Study on History State}
\label{app:aggre_mean}

We experiment with different choices of $h_i$ to investigate its impact on training. First, to experiment with other possible designs of $h_i$, we change the aggregation function from \textit{minimum} to \textit{mean}, in which case $h_i$ tracks the \textit{average} length of correct responses in the training history. The model receives a positive length reward whenever it generates a correct response that is shorter than the average of the previous correct ones. We denote this as \textbf{HAPO w/ $h_i= mean$}. Second, to ablate historical information from our length reward, we set $h_i$ to the \textit{average} length of correct responses in the \textit{current batch}. This effectively removes historical information and renders our method similar to Query-Opt \cite{arora2025}, except that we use our novel length reward. We denote this as \textbf{HAPO w/o $h_i$}. The results are shown in Table~\ref{tab:aggre_mean}.

\begin{table*}[h]
  \centering
  \resizebox{\textwidth}{!}{
  \begin{tabular}{lcccccccc}
    \toprule
    & \multicolumn{2}{c}{GSM8K} &  \multicolumn{2}{c}{MATH500} & \multicolumn{2}{c}{AIME2024}  & \multicolumn{2}{c}{Average} \\
    \cmidrule(r){2-9} 
    Models     & Pass@1 \textcolor{red}{$\uparrow$}  & \#Tokens \textcolor{red}{$\downarrow$} & Pass@1 \textcolor{red}{$\uparrow$}  & \#Tokens \textcolor{red}{$\downarrow$}  & Pass@1 \textcolor{red}{$\uparrow$}  & \#Tokens \textcolor{red}{$\downarrow$} & Pass@1 \textcolor{red}{$\uparrow$}  & \#Tokens \textcolor{red}{$\downarrow$} \\
    \midrule
    DeepSeek-R1-1.5B & 83.65  & 2066  & 79.15   & 6449  & 29.38  & 14615 & 64.06 & 7710 \\
    HAPO w/ $h_i= min$ & 79.08  & 661  & 81.05  & 2978  & 26.56  & 8171 & 62.23 & 3937 \\
    HAPO w/ $h_i= mean$ & 74.46  & 413 & 74.90  & 2676 & 18.33 & 7487 & 55.90 & 3523 \\
    HAPO w/o $h_i$ & 75.85 & 513 & 77.95 & 2754 & 24.60 & 11077 & 59.47 & 4781 \\
    \bottomrule
  \end{tabular}}
  \caption{\label{tab:aggre_mean} Results comparing different types of $h_i$.}
\end{table*}

\paragraph{HAPO w/ $h_i= mean$} Under this new update schedule, we expect $h_i$ to decrease more slowly during training, implying a slower reduction in response length. However, the results suggest otherwise. With the same number of training epochs, HAPO with the \textit{mean} update produces more concise responses but lower accuracy than HAPO with the \textit{minimum} update across the benchmarks. This indicates that HAPO tends to reduce response length more aggressively when using the historical mean. A possible explanation lies in the changes in reward distribution. When the \textit{minimum} function is used, the lengths of most newly sampled responses are likely to exceed $h_i$, making many length rewards concentrate in the negative flat region. This results in more cases with a reward variance of 0, which consequently gives weaker learning signals. In contrast, using the \textit{mean} function allows the sampled lengths to be more symmetrically distributed around $h_i$, thus leading to stronger learning signals.

\paragraph{HAPO w/o $h_i$} Compared with the original HAPO method, removing historical information increases the response length by a significant 36\% on AIME2024, while slightly reducing the length on GSM8K and MATH500. The accuracy decreases across the three benchmarks. These results show that historical information is essential for our method to balance accuracy and efficiency.

\section{Repetition Reward}

\begin{table*}[h]
  \centering
  \resizebox{\textwidth}{!}{
  \begin{tabular}{lcccccccc}
    \toprule
    & \multicolumn{2}{c}{GSM8K} &  \multicolumn{2}{c}{MATH500} & \multicolumn{2}{c}{AIME2024} & \multicolumn{2}{c}{Average} \\
    \cmidrule(r){2-9} 
    Models     & Pass@1 \textcolor{red}{$\uparrow$}  & \#Tokens \textcolor{red}{$\downarrow$} & Pass@1 \textcolor{red}{$\uparrow$}  & \#Tokens \textcolor{red}{$\downarrow$}  & Pass@1 \textcolor{red}{$\uparrow$}  & \#Tokens \textcolor{red}{$\downarrow$} & Pass@1 \textcolor{red}{$\uparrow$}  & \#Tokens \textcolor{red}{$\downarrow$} \\
    \midrule
    DeepSeek-R1-1.5B & 83.65  & 2066  & 79.15   & 6449  & 29.38  & 14615 & 64.06 & 7710 \\
    HAPO & 79.08  & 661  & 81.05  & 2978  & 26.56  & 8171 & 62.23 & 3937 \\
    ~~w/ Repetition Penalty & 73.57  & 404  & 73.05  & 2782  & 17.71  & 6996 & 54.78 & 3394 \\
    \bottomrule
  \end{tabular}}
  \caption{\label{tab:rep} Results with repetition reward.}
\end{table*}

Following \citet{yeo2025}, we incorporate an additional repetition penalty into the reward function. Specifically, we identify repeated n-grams within each generated response, count the total number of tokens that appear in these repeated n-grams, and divide this count by the total number of tokens in the response. The resulting repetition penalty, denoted as $rp$, is scaled by a weight constant $w_{\text{rp}}$ and subtracted from the overall reward. The final reward is given by:
\[
r_{i}^{(j)} = 1(a_{1}^{(j)} = a^*_i) + w \cdot rl_{i}^{(j)} - w_{\text{rp}} \cdot rp_{i}^{(j)}.
\]
In our experiments, we use 3-grams for repetition detection and set $w_{\text{rp}} = 1.0$. The results are presented in Table~\ref{tab:rep}.

Compared to HAPO, incorporating the repetition penalty results in shorter responses but leads to additional accuracy drops of approximately 5\%, 8\%, and 9\% on GSM8K, MATH500, and AIME2024, respectively. We suspect that the penalty may be too strong. Using larger n-grams and reducing the weight $w_{\text{rp}}$ could potentially improve the overall performance.

\section{Results with PPO}
\label{sec:ppo_results}

In addition to GRPO, we also experiment with PPO \cite{schulam2017ppo} as the training algorithm. We set the learning rate to 3e-6 and train the models for 5 epochs. The results obtained using PPO are presented in Table~\ref{tab:ppo_results}.

\begin{table*}
  \centering
  \resizebox{\textwidth}{!}{
  \begin{tabular}{lcccccccc}
    \toprule
    & \multicolumn{2}{c}{GSM8K} &  \multicolumn{2}{c}{MATH500} & \multicolumn{2}{c}{AIME2024} & \multicolumn{2}{c}{Average}                \\
    \cmidrule(r){2-9} 
    Models     & Pass@1 \textcolor{red}{$\uparrow$}  & \#Tokens \textcolor{red}{$\downarrow$} & Pass@1 \textcolor{red}{$\uparrow$}  & \#Tokens \textcolor{red}{$\downarrow$}  & Pass@1 \textcolor{red}{$\uparrow$}  & \#Tokens \textcolor{red}{$\downarrow$} & Pass@1 \textcolor{red}{$\uparrow$}  & \#Tokens \textcolor{red}{$\downarrow$} \\
    \midrule
    DeepScaleR-1.5B & \textbf{86.55}  & 1702  & \textbf{88.05}  & 3765  & \textbf{37.92}  & 8826  & \textbf{70.84} & 4764 \\
    HAPO (GRPO) & 83.59  & 1122  & 84.20  & \textbf{2681}  & 31.15  & \textbf{5813}  & 66.31 & \textbf{3205} \\
    ~~w/ PPO & 86.20 & \textbf{1032} & 87.00 & 2852 & 36.25 & 7548 & 69.82 & 3811 \\
    \midrule
    Qwen-2.5-1.5B-Inst & \textbf{70.43}  & 322  & \textbf{53.50} & 799  & \textbf{3.54}  & 3923 & \textbf{42.49} & 1681 \\
    HAPO (GRPO) & 67.90  & \textbf{200}  & 52.25  & \textbf{390}  & 2.81  & \textbf{1493} & 40.99 & \textbf{694} \\
    ~~w/ PPO & 63.74  & 448  & 41.45  & 724  & 0.83  &  3505 & 35.34 & 1559 \\
    \bottomrule
  \end{tabular}}
  \caption{\label{tab:ppo_results} PPO results on GSM8K, MATH500 and AIME2024.}
\end{table*}

As shown, on DeepScaleR-1.5B, PPO gives longer responses but higher accuracy than GRPO. Compared with the base model, PPO yields a 20\% reduction in response length. However, on Qwen-2.5-1.5B-Inst, PPO proves less effective, leading to a significant drop in accuracy with minimal reduction in length. Further investigation is warranted to understand the underlying causes of this discrepancy.

\section{Discussion}
\subsection{Limitations and Future Work}
\label{app:limitations}

There are several limitations to our work. First, we primarily focus on math problem solving for both training and evaluation, given its ease of automatic verification and its strong alignment with the reasoning capabilities of LLMs. Future work can apply HAPO to other reasoning tasks, such as multi-hop or scientific question answering, or potentially extend to more general-purpose settings. 

Second, due to limited computational resources, we did not conduct an extensive exploration of the length reward design. Our current approach uses a single scalar value aggregated from training history as the reference point, without leveraging other rich information that the history can provide. For example, one can track the full \textit{distribution} of past response lengths for each query and compute the length reward based on a percentile within that distribution.

\subsection{Broader Impact}
\label{app:impact}

HAPO can be applied to large reasoning models to substantially reduce token usage while preserving most of their original performance. This not only lowers computational costs (e.g., electricity consumption and carbon footprint), but also accelerates inference and reduces response latency—benefiting LLM service providers, end users, and the environment. 

While HAPO may introduce a slight decrease in accuracy, which can potentially increase the risk of hallucinations or misinformation, the inclusion of a correctness reward is designed to mitigate such degradation. We argue that this trade-off is carefully controlled and remains within an acceptable range.

\end{document}